\documentclass{article}
\usepackage{spconf,amsmath,graphicx}

\usepackage{color}

\usepackage{xspace}
\usepackage{amssymb}
\usepackage{multicol}
\usepackage{multirow}
\usepackage{cite}
\usepackage{caption}
\usepackage{subcaption}

\newcommand{\ie}{\emph{i.e.}\xspace}
\newcommand{\etal}{\emph{et al.}\xspace}
\newcommand{\eg}{\emph{e.g.}\xspace}

\newcommand{\mat}[1]{\mathbf{#1}}
\newcommand{\tr}{\mathsf{tr}}
\newcommand{\methodname}{DS-SBD\xspace}

\newlength\savewidth\newcommand\shline{\noalign{\global\savewidth\arrayrulewidth
  \global\arrayrulewidth 1.5pt}\hline\noalign{\global\arrayrulewidth\savewidth}}

\title{Unsupervised Video Anomaly Detection for Stereotypical Behaviours in Autism}
\name{Jiaqi Gao$^{1\ast}$ \quad Xinyang Jiang$^{2\dagger}$ \quad Yuqing Yang$^{2}$ \quad Dongsheng Li$^{2}$ \quad Lili Qiu$^{2}$\thanks{$^\ast$: Work was done during internship at MSRA.} \thanks{$^{\dagger}$: Corresponding author. Email: xinyangjiang@microsoft.com}}
\address{$^{1}$School of Computer Science, Fudan University \quad
$^{2}$Microsoft Research Asia}
\begin{document}
%
\maketitle
\begin{abstract}
Monitoring and analyzing stereotypical behaviours is important for early intervention and care taking in Autism Spectrum Disorder~(ASD). This paper focuses on automatically detecting stereotypical behaviours with computer vision techniques. Off-the-shelf methods tackle this task by supervised classification and activity recognition techniques. 
However, the unbounded types of stereotypical behaviours and the difficulty in collecting video recordings of ASD patients largely limit the feasibility of the existing supervised detection methods. 
As a result, we tackle these challenges from a new perspective, \ie unsupervised video anomaly detection for stereotypical behaviours detection. The models can be trained among unlabeled videos containing only normal behaviours and unknown types of abnormal behaviours can be detected during inference. Correspondingly, we propose a \textbf{D}ual \textbf{S}tream deep model for \textbf{S}tereotypical \textbf{B}ehaviours \textbf{D}etection, \textbf{\methodname}, based on the temporal trajectory of human poses and the repetition patterns of human actions. Extensive experiments are conducted to verify the effectiveness of our proposed method and suggest that it serves as a potential benchmark for future research.
\end{abstract}

\begin{figure*}[ht]
    \centering
    \includegraphics[width=0.75\linewidth]{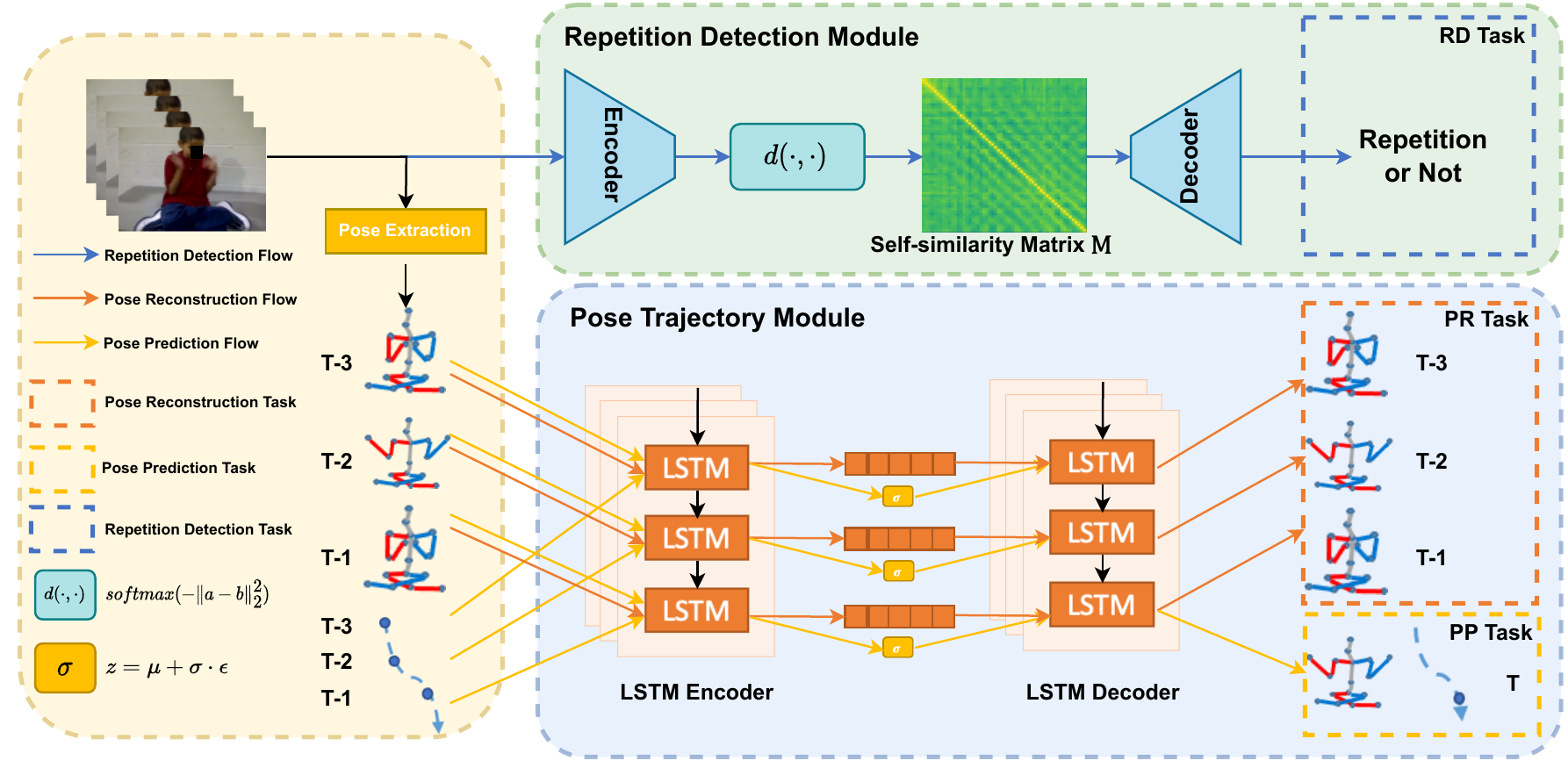}
    \caption{The overview of \methodname architectures. \methodname contains two task-specific modules, \ie repetition detection module and pose trajectory module, ensembled by three proxy tasks. PR: pose reconstruction, PP: pose prediction, RD: repetition detection. $\sigma$ represents the reparameterization trick used in the PP module.}
    \label{fig:model}
\end{figure*}
\vspace{-6pt}
\section{Introduction}
\label{sec:intro}

Autism spectrum disorder~(ASD) is a neurological and developmental disorder~\cite{rapin1991autistic}
that begins early in childhood and even lasts throughout a person's life. 
It causes problems with functioning in society and often affects how people interact, communicate, and socialize with others, resulting in stereotypical behaviours~\cite{rajagopalan2013self}. 
Stereotypical behaviours refer to the abnormal and non-functional repetitive behaviours that happens with no obvious stimulus, such as arm flapping, head banging, and spinning. 
They will negatively affect ASD children's performance on skill acquisition and social interaction, and as a stress indicator it could even lead to a meltdown event or cause self-damaging behaviours \cite{rad2018deep}. 
As a result, monitoring, evaluating, and analyzing the stereotypical behaviours are essential for the clinicians and caregivers to treat and take care of ASD patients, 
and an automated stereotypical behaviour detection system holds great potentials in the ASD patient caring and treatment. 

In this paper, we focus on automatically detecting stereotypical behaviours from video recording of the ASD patients.  
Ryan~\etal\cite{de2020computer} surveyed recent vision-based methods which mainly focus on how to correctly classify the stereotypical behaviours in autism with the help of action recognition~\cite{pandian2022detecting, pandey2020guided, wei2022vision} and video classification~\cite{lakkapragada2022classification, rajagopalan2014detecting, liang2021autism, sun2020spatial, anirudh2019bootstrapping} techniques. 
Existing methods perform well on a limited set of pre-defined stereotypical behaviour types through supervised learning paradigms. 
However, in practice, stereotypical behaviours detection is an open-set problem, where types of ASD stereotypical behaviours are unbounded with a large variance across different patients. 
Thus, there will always be novel behaviour types unseen in the training set, which previous methods are not able to detect. 
Furthermore, the collection of clinical video datasets brings great challenge, due to the privacy concerns and high data annotation cost from medical professionals.

To solve the challenge of unknown behaviour types and data collection difficulty, we propose to study ASD stereotypical behaviours detection from a new perspective, \ie unsupervised video anomaly detection (VAD). 
Unsupervised VAD learns the distribution of normal behaviours during training and  distinguishes the anomaly ASD behaviours as the outlier of the learned distribution. 
Since unsupervised VAD can detect any anomaly types out of the normal behaviours distribution, it is not limited by a finite set of pre-defined anomaly types. 
Furthermore, unsupervised VAD does not require to collect any data containing abnormal behaviours for training. Hence, it eases the burden of collecting clinic videos containing ASD patients. 
 
However, existing unsupervised VAD approaches~\cite{liu2021hybrid,chang2020clustering,liu2018future, gong2019memorizing,wang2022video} mainly focus on surveillance scenarios, and directly migrating them to stereotypical behaviour detection is non-trivial, for two reasons: 
1) Stereotypical behaviours of ASD patients contain a specific repetitive patterns, while exiting unsupervised VAD methods can not incorporate such prior knowledge. 
2) The videos of ASD patients are recorded under a unconstrained environment with various viewpoints and background noises, which brings challenges to the conventional unsupervised VAD methods focusing on surveillance videos under a constrained environment.  

As a result, we propose a novel \textbf{D}ual \textbf{S}tream network for \textbf{S}tereotypical \textbf{B}ehaviours \textbf{D}etection, \textbf{\methodname}, where each stream tackles one of the aforementioned two challenges respectively. 
Specifically, to improve the robustness over domain variance and background noises, we propose a pose trajectory module that models the temporal consistency of the human actions based on the temporal trajectory of human pose keypoints, filtering out the background noises and domain variance of the raw image frames.
Secondly, to incorporate the repetition pattern of ASD stereotypical behaviours, we propose a repetition detection module which detects the abnormal behaviours based on frame level repetitive patterns. 
The proposed \methodname is trained in an unsupervised fashion over videos containing only normal human behaviours with three proxy tasks, \ie pose reconstruction task, pose prediction task, and repetition detection task.

Our main contributions are summarized as follows: 
1) To tackle unknown behaviour types and data collecting difficulty, we formulate ASD stereotypical behaviours detection as an unsupervised video anomaly detection task and reorganize the existing self-stimulatory behaviour dataset~(SSBD) for evaluation. 
2) To leverage the ASD stereotypical behaviour prior knowledge and improve the robustness, we propose a dual stream abnormal detection network \methodname ensembled by novel pose trajectory and repetition detection modules.  
3) Extensive experimental results and ablation studies verify the effectiveness of each modules, suggesting \methodname could serve as a benchmark for this new challenging task in the future.

\section{Methodology}
\label{sec:method}
Fig. \ref{fig:model} shows the overall network structure of our proposed \methodname. It is a dual stream structure containing two modules, namely a pose trajectory module and a repetition prediction module. The pose trajectory module is responsible for detecting stereotypical behaviours based on human pose trajectories. The repetition module detects the abnormal behaviours based on the action repetitions over a certain period. Following the unsupervised video anomaly detection training settings, the training set only needs to contain videos with normal behaviours. 
The model is expected to learn the distribution of normal behaviours from training videos, and outputs a frame-level anomaly score at the inference time to judge whether it is an out of distribution abnormal behaviour. 
\subsection{Preliminaries}
Given a video with $N$ frames as $\mat{X}=[\mat{X}^1,\mat{X}^2, \ldots, \mat{X}^N]\in \mathbb{R}^{N \times C \times H \times W}$, the corresponding human poses with $K$ keypoints in $i$-th frame is denoted as $\mat{P}^i = [\mat{P}^i_1, \mat{P}^i_2, \ldots, \mat{P}^i_K] \in \mathbb{R}^{1 \times K \times d}$, where $d$ is the coordinate dimensions of human pose. The trajectory of $j$-th keypoints of one human pose is defined as $\tr(\mat{P}_j) = [\mat{P}^1_j, \mat{P}^2_j, \ldots, \mat{P}^N_j] \in \mathbb{R}^{N \times 1 \times d}$. 

\subsection{Pose Trajectory Module}
The pose trajectory module is trained with two proxy tasks, \ie the pose reconstruction task and the pose prediction task. 

\subsubsection{Pose Reconstruction}
The pose reconstruction (PR) proxy task takes the assumption that the pose trajectories of normal behaviours can be well reconstructed by an autoencoder while the anomaly behaviours can not. 
Specifically, an LSTM based autoencoder $\mathcal{F}^{\mathrm{PR}}$ is proposed for the reconstruction proxy task. 
$\mathcal{F}^{\mathrm{PR}}$ takes a human pose trajectory $\tr(\mat{P})$ as input and aims at reconstructing each human pose keypoints in this trajectory during training. A MSE training loss $\mathcal{L}^{\mathrm{PR}}$ is used to optimize $\mathcal{F}^{\mathrm{PR}}$: 
\begin{equation}
\begin{split}
    \mathcal{L}^{\mathrm{PR}} &= \| \mathcal{F}^{\mathrm{PR}}(
    \tr(\mat{P})) - \tr(\mat{P}) \|^2_2 \\
    &= \sum_{i=1}^N\sum_{j=1}^K\| \mathcal{F}^{\mathrm{PR}}(
    \mat{P}^i_j) - \mat{P}^i_j \|^2_2
\end{split}
\end{equation}

During inference, the pose reconstruction errors of $\mathcal{F}^{\mathrm{PR}}$ on each keypoint in a frame is summed up the to get a frame-level anomaly score:
\begin{equation}
    s^{\mathrm{PR}}_i = \sum_{j=1}^K\| \mathcal{F}^{\mathrm{PR}}(
    \mat{P}^i_j) - \mat{P}^i_j \|^2_2
\end{equation}

\subsubsection{Pose Prediction}
The pose prediction (PP) proxy task assumes that 
normal human behaviours are temporally consistent, while abnormal ones usually come with unexpected change of actions.
Specifically, given a trajectory of $T$ consecutive poses $\tr(\mat{P}^{1:T})$, the pose prediction task attempts to forecast the  next human pose $\mat{P}^{T+1}$ with a deep model $\mathcal{F}^{\mathrm{PP}}$:
\begin{equation}
    \hat{\mat{P}}^{T+1} = \mathcal{F}^{\mathrm{PP}}(
    \tr(\mat{P}^{1:T}))
\end{equation}

Following~\cite{parsaeifard2021learning}, the pose prediction task is built upon local pose trajectory~(all keypoints) and global pose trajectory~(center point of all keypoints) forecasting. For local keypoints, $\mathcal{F}^{\mathrm{PP}}$ is an LSTM-based variational autoencoder. For a center point, a cascaded LSTM is used for prediction. Similar to the pose reconstruction task, we use MSE to optimize $\mathcal{F}^{\mathrm{PP}}$:
\begin{equation}
\begin{split}
    \mathcal{L}^{\mathrm{PP}} &= \| \hat{\mat{P}}^{T+1}_c - \mat{P}^{T+1}_c \|^2_2 + \| \hat{\mat{P}}^{T+1} - \mat{P}^{T+1} \|^2_2 \\
    &= \| \hat{\mat{P}}^{T+1}_c - \mat{P}^{T+1}_c \|^2_2 + \sum_{j=1}^K\| \hat{\mat{P}}^{T+1}_j - \mat{P}^{T+1}_j \|^2_2 
\end{split}
\end{equation}
where $\hat{\mat{P}}^{T+1}$ and $\hat{\mat{P}}_c^{T+1}$ are the predicted local pose keypoints and the global pose keypoint of the $T+1$ frame, respectively. 

Similar to the pose reconstruction module, the anomaly score of one frame is its forecasting errors given past trajectories: 
\begin{equation}
    s^{\mathrm{PP}}_i = \| \hat{\mat{P}}^i_c - \mat{P}^i_c \|^2_2 + \sum_{j=1}^K\| \hat{\mat{P}}^i_j - \mat{P}^i_j \|^2_2 
\end{equation}

\subsection{Repetition Detection Module}
We observe that one of the most distinct characteristics of the stereotypical behaviours in autism spectrum disorder is the repetitive pattern. In other words, the anomaly behaviours would be repeated periodically over short time intervals in the videos. 
To leverage this essential prior knowledge, we propose a repetition detection module~(RD). Inspired by recent repetition counting methods~\cite{levy2015live, dwibedi2020counting, hu2022transrac}, we model the repetitive patterns as a temporal self-similarity matrix $\mat{M}$, 
whose elements $\mat{M}_{i,j}$ are the similarity score between the feature embedding of $i$-th frame and $j$-th frame, followed by the row-wise softmax operation~\cite{dwibedi2020counting}.
\begin{equation}
    \mat{M}_{i,j} = \mathrm{softmax}( - \| x_i - x_j \|^2_2 )
\end{equation}
where $x_i$ and $x_j$ are the latent feature embeddings of $i$-th and $j$-th frames.

Based on the self-similarity matrix, the repetition detection module $\mathcal{F}^{\mathrm{RD}}$ 
outputs an anomaly score for each video frame, showing the probability whether this frame contains repetitive actions: 
\begin{equation}
    s^{\mathrm{RD}}_i = \mathcal{F}^{\mathrm{RD}}(\mat{X}^i)
\end{equation}
where $s^{\mathrm{RD}}_i$ is the $i$-th frame anomaly score of the repetition detection module, and $\mat{X}^i$ is the $i$-th input frame. The proposed repetition detection module can be trained on public repetition counting dataset or videos synthesized from the VAD training set.

\subsection{Anomaly Score}
The ultimate anomaly score for each video frame $S_i$ is the weighted sum of two anomaly scores from the pose trajectory module and one anomaly score from the repetition detection module:
\begin{equation}
    S_i = \alpha \cdot \frac{s^{\mathrm{PR}}_i - \mu_{\mathrm{PR}}}{\sigma_{\mathrm{PR}}} + \beta \cdot \frac{s^{\mathrm{PP}}_i-\mu_{\mathrm{PP}}}{\sigma_{\mathrm{PP}}} + \gamma \cdot s^{\mathrm{RD}}_i 
\end{equation}
where $\alpha$, $\beta$, and $\gamma$ are the weights of the three anomaly scores, $\mu_{\mathrm{PR}}$, $\sigma_{\mathrm{PR}}$, $\mu_{\mathrm{PP}}$, and $\sigma_{\mathrm{PP}}$ are the means and standard deviations of training pose reconstruction and prediction errors.

\section{Experiments}
\label{sec:exp}

\subsection{Dataset}
In our experiments, we use the self-stimulatory behaviour dataset (SSBD)~\cite{rajagopalan2013self} to evaluate the models, which is the publicly-available benchmarking dataset for stereotypical behaviour detection. 
The SSBD dataset contains 75 videos with three stereotypical behaviours, \ie arm flapping, head banging, and spinning. 
Following the setting of unsupervised VAD, we split the dataset to testing set with 20 videos and training set with rest of the videos. All sub-clips containing stereotypical behaviours are excluded in the training videos. 

\subsection{Implementation Details}
We choose the Adam optimizer for training and the learning rate is set to 0.004. AlphaPose~\cite{fang2017rmpe} is used to generate the 2D human pose and VideoPose3D~\cite{videopose3d2019} is used to generate the 3D human pose. The batch size is set to 60 and the number of consecutive frames in one batch $T$ is set to 64. The repetition detection module applies the backbone of RepNet~\cite{dwibedi2020counting}. 
We provide the non-overlapping sliding windows of $T$ frames to compute the final frame-level anomaly scores during testing.
\vspace{-10pt}

\subsection{Results}
We use two widely used evaluation metrics in video anomaly detection community, \ie micro-averaged area under receiver operation characteristic curve~(AUROC), and macro-averaged AUROC, to evaluate the models. Specifically, the micro-averaged AUROC is to compute the overall frame-level AUC by concatenating all the frames during testing. The macro-averaged AUROC is the average of AUC grouped by videos varying the threshold. 

\begin{figure}[h]
    \centering
    \includegraphics[width=\linewidth, trim=5 8 0 0, clip]{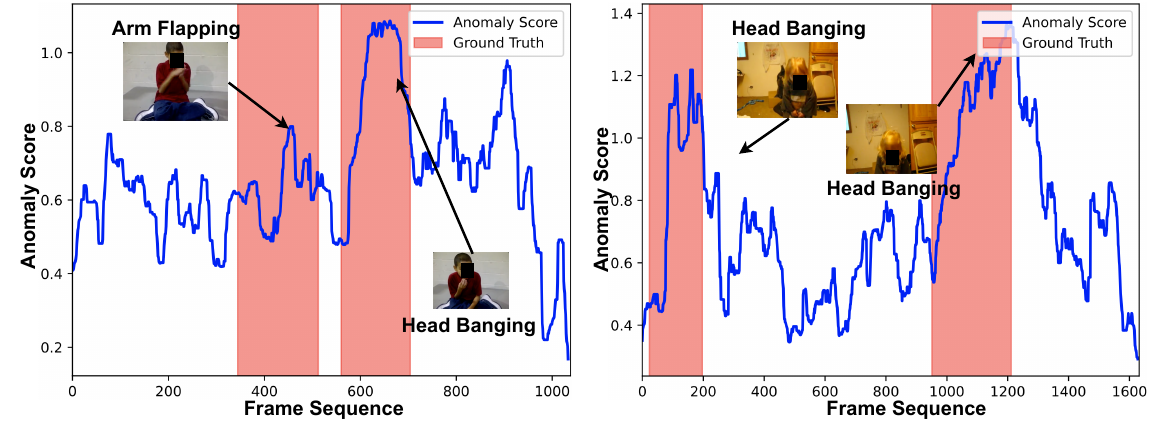}
    \caption{The visualization results of our proposed method. }
    \label{fig:visualization}
\end{figure}

\begin{table}[h]
\centering
\vspace{-10pt}
\caption{The quantitative comparison results between the state-of-the-art model and our proposed model.}
\label{tab: result1}
\vspace{-5pt}
\begin{tabular}{lcc}
\shline
\multirow{2}{*}{\textbf{Method}}   & \multicolumn{2}{c}{\textbf{AUROC}} \\ \cline{2-3}
& \textbf{micro}   & \textbf{macro} \\ \hline
Frame-Pred.~\cite{liu2018future}  & 52.52\% & 54.93\%  \\
MNAD~\cite{park2020learning}  & 53.70\% & 56.45\%  \\
HF2VAD~\cite{liu2021hybrid}  & 60.43\% & 54.35\%  \\
\methodname-PR & 54.54\% & 51.88\%   \\
\methodname-PP & 62.01\% & 55.54\%   \\
\methodname-RD & 69.87\% & 72.81\%   \\
\methodname     & \textbf{71.04\%} & \textbf{73.39\%}   \\
\shline
\end{tabular}
\vspace{-8pt}
\end{table}
We report the results of our \methodname with different proxy tasks and compare with several state-of-the-art unsupervised VAD methods, including Frame-Pred.~\cite{liu2018future}, HF2VAD~\cite{liu2021hybrid}, and MNAD~\cite{park2020learning} in Table~\ref{tab: result1}.  The model performance is boosted with the three effective auxiliary tasks from 54.54\% to 71.04\% of micro-AUROC and the best macro-AUROC reaches 73.39\%, which significantly outperforms the baseline models. In addition, we observe that the repetition detection module plays a dominant role of unsupervised video anomaly detection for autism spectrum disorder because the stereotypical behaviours are often characterized by repetition. The visualization results are shown in Fig~\ref{fig:visualization}. 

\vspace{-25pt}
\subsection{Ablation Study}
\label{sec: ablation}
We conduct ablation studies to investigate the factors that may contribute to the anomaly detection performance. 

\begin{table}[ht]
\centering
\caption{The ablation study of different pose modalities.}
\label{tab: result2}
\vspace{-5pt}
\resizebox{0.9\linewidth}{!}{
\begin{tabular}{ccccccc}
\shline
\multicolumn{2}{c}{\textbf{PR}} & \multicolumn{2}{c}{\textbf{PP}} & \multicolumn{1}{c}{\multirow{2}{*}{\textbf{RD}}} & \multicolumn{2}{c}{\textbf{AUROC}} \\ \cline{6-7}
\textbf{2D}    & \textbf{3D}    & \textbf{2D}    & \textbf{3D}    & \multicolumn{1}{c}{}                     & \textbf{micro}   & \textbf{macro}                       \\ \hline
\checkmark & & & & & 54.54\% & 51.88\%                              \\
           & \checkmark & & & & 57.85\% & 61.75\%                   \\
           & & \checkmark & & & 62.01\% & 55.54\%                   \\
           & & & \checkmark & & 60.34\% & 61.40\%                   \\
           & & & & \checkmark & 69.87\% & 72.81\%                   \\
\checkmark & & \checkmark & & & 61.99\% & 55.53\%                   \\
\checkmark & & \checkmark & & \checkmark & \textbf{71.04\%} & \textbf{73.39\%}\\
           & \checkmark & & \checkmark & & 60.30\% & 61.42\%        \\
           & \checkmark & & \checkmark & \checkmark & 70.65\% & 73.32\% \\ \shline
\end{tabular}
}
\vspace{-15pt}
\end{table}
\noindent\textbf{2D pose vs. 3D pose.}
Although, 3D skeleton trajectory can provide the depth information of human motion, it is usually not as stable and robust compared to 2D pose prediction models because inferring 3D information from 2D frames is more challenging. 
As shown in Table~\ref{tab: result2}, our method achieves better performance when taking 2D poses as input. 

\noindent\textbf{Number of frames.}
Considering the temporal consistency and periodicity of each stereotypical behaviours, we also investigate whether the different number of input frames will affect the performance. As shown in Table~\ref{tab: result3}, the model achieves the best performance when the input is a relatively long sequence of frames~(\eg $T=64$). This is because stereotypical behaviours with low frequency often require more information from history frames to accurately discover a periodic repetition patterns.  

\noindent\textbf{Weight estimation.} We estimate the $\alpha$, $\beta$, and $\gamma$ by grid search from 0 to 3. In Table~\ref{tab: result4}, the DB-SBD* achieves the best performance when $\alpha$=1.5, $\beta$=0.2, $\gamma$=1.3 with the marginal improvement compared with the default settings~($\alpha$=$\beta$=$\gamma$=1), which shows our model is relatively robust.

\begin{table}[h]
\vspace{-10pt}
    \begin{minipage}[h]{0.23\textwidth}
        \centering
        \caption{The ablation study of different input frames.}
        \label{tab: result3}
        \vspace{-5pt}
        \resizebox{\linewidth}{!}{
            \begin{tabular}{ccc}
            \shline
            \multicolumn{1}{c}{\multirow{2}{*}{$T$ \textbf{frames}}} & \multicolumn{2}{c}{\textbf{AUROC}} \\ \cline{2-3} 
            \multicolumn{1}{c}{}                        & \textbf{micro}   & \textbf{macro}   \\ \hline
            4                                           & 69.37\%     & 72.74\%     \\
            8                                           & 70.07\%     & 72.34\%     \\
            16                                          & 70.12\%     & 72.93\%     \\
            64                                          & \textbf{71.04\%}     & \textbf{73.39\%}     \\ \shline
            \end{tabular}
        }
    \end{minipage}
    \begin{minipage}[h]{0.23\textwidth}
        \centering
        \caption{The ablation study of weight estimations.}
        \label{tab: result4}
        \vspace{-5pt}
        \resizebox{\linewidth}{!}{
            \begin{tabular}{lcc}
            \shline
            \multicolumn{1}{c}{\multirow{2}{*}{\textbf{Method}}} & \multicolumn{2}{c}{\textbf{AUROC}} \\ \cline{2-3} 
            \multicolumn{1}{c}{}                        & \textbf{micro}   & \textbf{macro}   \\ \hline
            DS-SBD                                          & {71.04\%}     & {73.39\%}     \\
            DS-SBD*                                          & \textbf{71.77\%}     & \textbf{73.58\%}     \\ \shline
            \end{tabular}
        }
     \end{minipage}
\vspace{-15pt}
\end{table}
\section{Conclusion}
\label{sec:conclusion}

In this paper, we drive a new research perspective of stereotypical behaviours detection in autism spectrum disorder, \ie unsupervised video anomaly detection. To better leverage the prior knowledge of ASD and improve the robustness, we propose a dual stream deep model \methodname that detects abnormal behaviours based on temporal trajectory of human poses and the repetition patterns of human actions. 
Extensive experimental results demonstrate the effectiveness of our method and may act as a benchmark for future research. In the future, we will investigate more simple but effective proxy tasks to boost the model discriminability.

\vfill\pagebreak

\bibliographystyle{IEEEbib}
\bibliography{refs}

\end{document}